\def\year{2020}\relax
\documentclass[letterpaper]{article} 
\usepackage{aaai20}  
\usepackage{times}  
\usepackage{helvet} 
\usepackage{courier}  
\usepackage[hyphens]{url}  
\usepackage{graphicx} 
\usepackage{graphicx}  
\usepackage{times}
\usepackage{latexsym}
\usepackage{soul}
\usepackage{url}
\usepackage{enumitem}
\setitemize{noitemsep,topsep=0pt,parsep=0pt,partopsep=0pt}

\usepackage{amsmath}
\usepackage{graphicx}
\usepackage{amsfonts}

 \usepackage{multirow}
 \usepackage[normalem]{ulem}
 \useunder{\uline}{\ul}{}
 \usepackage{booktabs} 
\usepackage{xcolor}
\usepackage{footnote}
\makesavenoteenv{tabular}

\usepackage{algorithm}
\usepackage{algpseudocode}

\algnewcommand{\algorithmicand}{\textbf{ and }}
\algnewcommand{\algorithmicor}{\textbf{ or }}
\algnewcommand{\OR}{\algorithmicor}
\algnewcommand{\AND}{\algorithmicand}

\newcommand{\citet}[1]{\citeauthor{#1}~\shortcite{#1}}
\newcommand{\citep}{\cite}

\title{\textsc{Hal}: Improved Text-Image Matching by Mitigating Visual Semantic Hubs }

\author{\Large \textbf{Fangyu Liu\textsuperscript{\rm 1}\thanks{Equal contributions.}\thanks{Correspondence to F. Liu \texttt{$<$fl399@cam.ac.uk$>$}.}, Rongtian Ye\textsuperscript{\rm 2}$^\ast$, Xun Wang\textsuperscript{\rm 3}$^\ast$}, Shuaipeng Li\textsuperscript{\rm 4}\\ 
\textsuperscript{\rm 1}University of Cambridge, Cambridge, UK\\ 
\textsuperscript{\rm 2}Aalto University, Espoo, Finland\\ 
\textsuperscript{\rm 3}Malong Technologies, Shenzhen, China\\
\textsuperscript{\rm 4}SenseTime Research, Beijing, China\\}

\begin{document}

\maketitle

\begin{abstract}
The \emph{hubness problem} widely exists in high-dimensional embedding space and is a fundamental source of error for cross-modal matching tasks. In this work, we study the emergence of hubs in \emph{Visual Semantic Embeddings} (VSE) with application to text-image matching. We analyze the pros and cons of two widely adopted optimization objectives for training VSE and propose a novel hubness-aware loss function (\textsc{Hal}) that addresses previous methods' defects. Unlike \cite{faghri2018vse++} which simply takes the hardest sample within a mini-batch, \textsc{Hal} takes all samples into account, using both local and global statistics to scale up the weights of ``hubs''. We experiment our method with various configurations of model architectures and datasets. The method exhibits exceptionally good robustness and brings consistent improvement on the task of text-image matching across all settings. Specifically, under the same model architectures as \cite{faghri2018vse++} and \cite{lee2018stacked}, by switching only the learning objective, we report a maximum R@1 improvement of 7.4\% on MS-COCO and 8.3\% on Flickr30k.\footnote{Our code is released at: \url{https://github.com/hardyqr/HAL}.}

\end{abstract}

\section{Introduction}
The hubness problem is a general phenomenon in high-dimensional space where a small set of source vectors, dubbed hubs, appear too frequently in the neighborhood of target vectors \cite{radovanovic2010hubs}. As embedding learning goes deeper, it has been a concern in various contexts including object classification \cite{tomavsev2011influence}, image feature matching \cite{jegou2008accurate} in Computer Vision and word embedding evaluation \cite{schnabel2015evaluation,faruqui2016problems}, word translation \cite{dinu2015improving,lazaridou2015hubness} in NLP. It is described as ``a new aspect of the dimensionality curse'' \cite{bellman1961adaptive,schnitzer2012local}.

In this work, we study the hubness problem in the task of text-image matching. In recent years, deep neural models have gained a significant edge over non-neural methods in cross-modal matching tasks~\cite{wang2016comprehensive}. Text-image matching has been one of the most popular ones among them. Most deep methods involve two phases: 1) training: two neural encoders (one for image and one for text) are learned end-to-end, mapping texts and images into a joint space, where items (either texts or images) with similar meanings are close to each other; 2) inference: for a query vector in modality A, a nearest neighbor search is performed to match the query vector against all item vectors in modality B. As the embedding space is learned through jointly modeling vision and language, it is often referred as \emph{Visual Semantic Embeddings} (VSE). Recent work on VSE has shown a clear trend of growing dimensions in order to obtain better embedding quality \cite{wehrmann2018bidirectional}. With deeper embeddings, visual semantic hubs increase dramatically. Such property is undesired as the data is structured in the form of text-image pairs and a one-to-one mapping firmly exists among all text and image points.
\begin{figure}
    \centering
    \includegraphics[width=8cm]{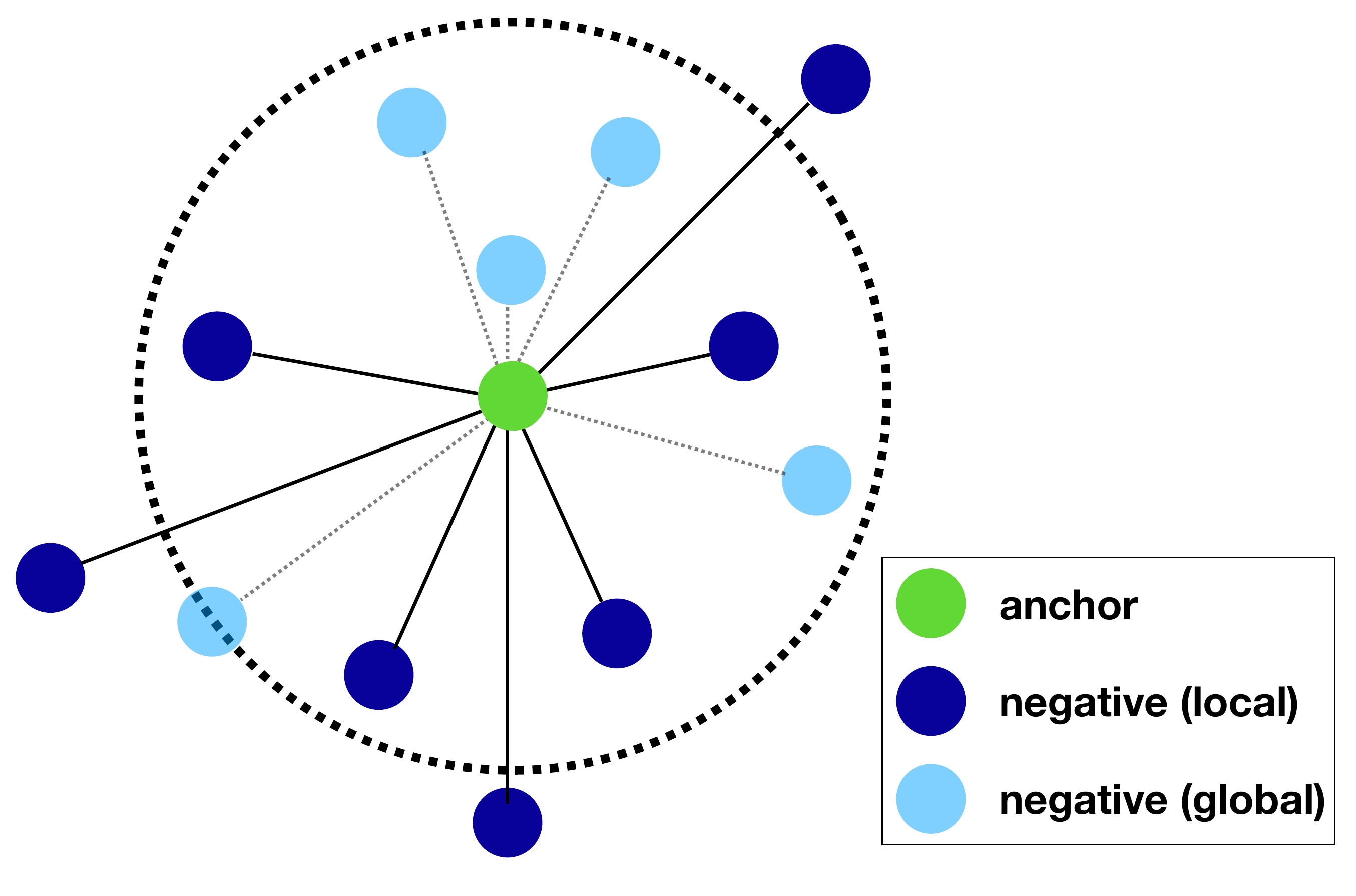}
    \caption{Visualization of our proposed objective, which is to leverage both local and global negative samples to identify hubs in high-dimensional embeddings and learn to avoid them. Local negatives are the ones within mini-batch while global ones are sampled from the whole training set.}
    \label{fig:my_label}
\end{figure}

However, the hubness problem is neither well noticed nor well addressed by current methods of training VSE. Since the start of this line of work \cite{frome2013devise,kiros2014unifying}, VSE models use either sum-margin (\textsc{Sum}, Eq.~\eqref{eq:sum-margin}) or max-margin (\textsc{Max}, Eq.~\eqref{eq:max-margin}) ranking loss (both are triplet based) to cluster the positive pairs and push away the negative pairs. \textsc{Sum} is robust across various settings but treats all triplets equivalently and utilizes no information from hard samples, thus does not address the hubness problem at all. \textsc{Max} excels at mining hard samples and achieved state-of-the-art on MS-COCO \cite{faghri2018vse++}. However, it does not explicitly consider the hubness problem, nor does it resist noise well. New models on training VSE have been consistently brought up in recent years. They include incorporating extra knowledge to augment original data, eg. generating adversarial samples \cite{shi2018learning}, and designing high-level objective that utilizes pre-trained models to align salient entities across modalities \cite{lee2018stacked,wu2019unified}. However, ever since \cite{faghri2018vse++}, the basic scheme of training VSE has not been enhanced. In this work we show that exploiting the data per se has yet reached its limit.

To fully extract the information buried within, we combine robustness with hard sample mining, proposing a self-adjustable hubness-aware loss called \textsc{Hal}. \textsc{Hal} takes both global (sampled from the whole training set) and local statistics (obtained from mini-batch) into account, leveraging information of hubs to automatically adjust weights of samples. It learns from hard samples and is robust to noise at the same time by taking multiple samples into account. Specifically, we exploit a sample's relationship to 1) other samples within the mini-batch; 2) its $k$-nearest neighbor queries in a memory bank, to decide its weight. The larger a hub is, the more it should contribute to the loss, resulting in a mitigation of hubs and an improvement of embedding quality. Through a thorough empirical comparison, we show that our method outperforms \textsc{Sum} and \textsc{Max} loss on various datasets and architectures by large margins. 

The major contribution of this work is a novel training objective (\textsc{Hal}) that utilizes both local and global statistics to identify hubs in high-dimensional embeddings. Compared with strong baselines \cite{faghri2018vse++} and \cite{lee2018stacked}, \textsc{Hal} improves \textsf{R@}$1$ by a maximum of $7.4\%$ on MS-COCO and $8.3\%$ on Flickr30k.

\section{Method}
We first introduce the basic formulation of VSE model; then review widely-adopted methods that we will compare to; in the end, propose our intended loss function. 

\subsection{Basic Formulation}
\label{sec:bf}
The bidirectional text-image matching framework consists of a text encoder and an image encoder. The text encoder is composed of word embeddings, a GRU~\cite{chung2014empirical} (or other sequential models) layer and a temporal pooling layer. The image encoder is usually a deep CNN and a linear layer. We use ResNet152~\cite{he2016deep}, Inception-ResNet-v2 (IRv2)~\cite{szegedy2017inception} and VGG19 \cite{simonyan2014very} pre-trained on ImageNet~\cite{deng2009imagenet} in our models. We denote them as functions $f$ and $g$, which map text and image to some vectors of size $d$ respectively.

For a text-image pair $(t,i)$, the similarity of $t$ and $i$ is measured by cosine similarity:
\begin{equation}
S_{it} = \bigg\langle  \frac{f (t)}{\| f (t) \|_2}, \frac{g (i)}{\| g (i) \|_2} \bigg\rangle : \mathbb{R}^{d} \times \mathbb{R}^{d}\rightarrow \mathbb{R}.
\end{equation}

During training, a margin based triplet ranking loss is usually adopted to cluster positive pairs and push negative pairs away from each other. There are mainly two prevalent choices which are \textsc{Sum} and \textsc{Max}. We introduce them in the next section along with our newly proposed non-triplet-based loss \textsc{Hal}.

\subsection{Revisit Two Triplet-based Loss Functions}
\label{sec:TO}
In this section we review the two popular loss functions that have been adopted for training VSE and analyze their pros and cons. 

\subsubsection{Sum-margin Loss (\textsc{Sum}).}
\label{sec:sum-margin}
\textsc{Sum} is a standard triplet loss adopted from the metric learning literature and has been used for training VSE since the start of this line of work \cite{frome2013devise,kiros2014unifying}. Its early form can be found in \cite{weston2010large} which was used for training joint word-image embeddings. Formally, \textsc{Sum} is defined as:
\begin{equation}
\begin{split}
 \mathcal{L_{\textsc{Sum}}} = \sum_{i\in I} \sum_{\bar{t}\in T\backslash \{t\}}  [\ \alpha -S_{it}+S_{i\bar{t}}\ ]_+ \\ 
+ \sum_{t\in T} \sum_{\bar{i}\in I\backslash \{i\}} [\ \alpha -S_{ti}+S_{t\bar{i}}\ ]_+,
\end{split}
\label{eq:sum-margin}
\end{equation}
where $[\cdot]_+ = \max(0,\cdot)$; $\alpha$ is a preset margin; $T$ and $I$ are text and image encodings in a mini-batch; $t$ is the descriptive text for image $i$ and vice versa; $\bar{t}$ denotes non-descriptive texts for $i$ while $\bar{i}$ denotes non-descriptive images for $t$.

The major shortcoming of \textsc{Sum} lies in the fact that it views all valid triplets within a mini-batch as equal and assigns identical weights to all, leading to a failure of identifying informative pairs. As we will detail in the following, a simple ``hard'' weighting by taking only the hardest triplet can greatly enhance a triplet-based loss's performance in training VSE.

\subsubsection{Max-margin Loss (\textsc{Max}).}
\label{sec:max-margin}
\citet{faghri2018vse++} proposed \textsc{Max} fairly recently (2018). \textsc{Max} differs from \textsc{Sum} by considering only the largest violation of margin within the mini-batch instead of summing over all margins:
\begin{equation}
\begin{split}
\mathcal{L_\textsc{Max}} = \sum_{i\in I} \max_{\bar{t}\in T\backslash \{t\}}  [\ \alpha -S_{it}+S_{i\bar{t}}\ ]_+ \\ 
+ \sum_{t\in T} \max_{\bar{i}\in I\backslash \{i\}} [\ \alpha -S_{ti}+S_{t\bar{i}}
\ ]_+.
\end{split}
\label{eq:max-margin}
\end{equation}

We refer to \textsc{Max} as a ``hard'' weighting strategy as it implicitly assigns a weight of $1$ to the hardest triplet and $0$ to all other triplets. Though \textsc{Max} was not used in the context of VSE before, it was thoroughly exploited in other embedding learning tasks \cite{wu2017sampling}. As analyzed by \cite{wu2017sampling}, a rigid stress on hard negatives like \textsc{Max} makes its gradient easily dominated by noise, being a result of either deficiency of the model architecture or data's structure per se. Through error analysis, we notice that the existence of \emph{pseudo hardest negatives} in training data is a major source of noise for \textsc{Max}. During training, only the hardest negative in a mini-batch is considered. If that sample contained happens to be incorrectly labeled or inaccurate, misleading gradients would be imposed on the network. Notice that \textsc{Sum} eases such noise in labels by taking all mini-batch's samples into account. When a small set of samples are with false labels, their false gradients would be canceled out by other correct negatives within the mini-batch, preventing the model from an optimization failure or overfitting to incorrect labels. That being said, \textsc{Sum} fails to make use of hard samples and does not address the hubness problem at all. It thus performs poorly on a well-labeled dataset like MS-COCO. 

Besides, both \textsc{Sum} and \textsc{Max} are triplet based, considering only one positive pair and one negative pair at a time. Such sampling manner isolates each triplet and disregards the overall distribution of data points. What's more, the triplet-style heuristics is easy for selected triplets to satisfy after the early stage of training, leaving very little information in gradients in the late stage \cite{yu2018hard}. As opposed to triplet loss, our proposed NCA-based loss, to be introduced in the next section, characterizes the whole local neighborhood and take the affinities among all pairs into consideration.

\subsection{The Hubness-Aware Loss ({\textsc{Hal}})}
\label{sec:HAL}
On the one hand, we obtain the greatest possible robustness through considering multiple samples; on the other hand, we try to make sure the samples being considered are hard enough - so that the training is effective. We tackle this problem by leveraging information from visual semantic hubs. Inspired by Neighborhood Component Analysis (NCA) \cite{goldberger2005neighbourhood} used for classification task, we propose a self-adaptive Hubness-Aware Loss (\textsc{Hal}) that weights samples within a mini-batch according to both local and global statistics. More specifically, \textsc{Hal} assigns more weights to samples which appear to be hubs (being close neighbors to multiple queries), judging from both the current mini-batch and a memory bank sampled from the whole training set. 

How global and local information are used will be detailed shortly. Before that, we briefly explain NCA and discuss why it is a natural choice for addressing hubness problem. In the classification context, NCA is formulated as:

\begin{equation}
\begin{split}
\mathcal{L_\textsc{Nca}} = \sum_{i=1}^{N}\bigg(\log \sum_{y_i = y_j}e^{S_{ij}} - \log \sum_{k=1}^{N} e^{S_{ik}}\bigg),
\end{split}
\end{equation}
where $N$ is the number of samples. And the gradient of $\mathcal{L_\textsc{Nca}} $ w.r.t.
positive and negative samples are computed as:
\begin{equation}
\begin{split}
w^+ =\bigg| \frac{\partial \mathcal{L_{\textsc{Nca}}}}{\partial {S_{ij}}^+} \bigg| = \frac{e^{S_{ij}}}{\sum_{y_i = y_j}e^{S_{ij}}} - \frac{e^{S_{ij}}}{\sum_{k=1}^N e^{S_{ik}}},\\
w^- = \bigg|\frac{\partial \mathcal{L_{\textsc{Nca}}}}{\partial {S_{ij}}^-}\bigg|  = \frac{e^{S_{ij}}}{\sum_{k=1}^N e^{S_{ik}}}.
\end{split}
\end{equation}
For a sample $S_{ij}$, when it is a close neighbor to multiple items in the search space, ie. being a hub, its weight as a positive is reduced and that as a negative is scaled up, meaning that it receives more attention during training. This basic philosophy of NCA will be used in both the local and global weighting schemes in the following.


\subsubsection{a) Global weighting through Memory Bank (\textsc{Mb}).}
One of the most desired property of an NCA-based loss is that it automatically assigns weights to all samples in one batch of back-propagation through computing gradients as suggested above. The more data points we have, the more reliable a hub can be identified. The most ideal approach of leveraging hubs is utilizing the idea of NCA and searching for hubs across the whole training set, so that all samples are compared against each other and information is made fully use of. However, it is computationally infeasible to minimize such objective function on a global scale - especially when it comes to computing gradients for all training samples \cite{wu2018improving}. We thus design hand-crafted criteria that follows the NCA's idea to explicitly compute weight of samples but does not require gradient computation. Specifically, at the beginning of each epoch, we sample all over training set and compute their embeddings to create a memory bank $M$ that approximates the global distribution of training data. Then we utilize relationships among mini-batch and memory bank to compute a global weight for each sample in the batch, highlighting hubs and passing the weight to the next stage of local weighting.

We define a function $\texttt{kNN}(x,M,k)$ to return the $k$ closest points (measured by $l_2$ distance) in point set $M$ to $x$ and the global weighting of {\textsc{Hal}} can be formulated as:
\begin{equation}
\small
\begin{split}
    W_{ii} =  1 -  e^{\alpha (S_{ii} -\epsilon_1)} / \\ \bigg(  e^{\alpha (S_{ii}-\epsilon_1)} + \sum_{\bar{t}\in K_1}e^{\alpha (S_{i\bar{t}} - \epsilon_2)} + \sum_{\bar{j}\in K_2}e^{\alpha (S_{\bar{j}i}-\epsilon_2)} \bigg)  ,\\
W_{it} =  \bigg(\ \sum_{\bar{t}\in K_1}e^{\beta( S_{i\bar{t}}-\epsilon_2)} + \sum_{\bar{i}\in K_2}e^{\beta (S_{\bar{i}t}-\epsilon_2)} \bigg) / \\
\bigg( e^{\beta (S_{ii}-\epsilon_1)}+ e^{\beta (S_{tt}-\epsilon_1)} +  \sum_{\bar{t}\in K_1}e^{\beta (S_{i\bar{t}}-\epsilon_2)} +\sum_{\bar{i}\in K_2}e^{\beta (S_{\bar{i}t}-\epsilon_2)} \bigg),
\end{split}
\label{eq:local_scaling_loss_new}
\end{equation}
where $W_{ii},W_{it}$ represent weight of positive and negative samples respectively; $K_1 = \texttt{kNN}(i,M_T\backslash\{t\},k), K_2 = \texttt{kNN}(t,M_I\backslash\{i\},k)$; $\alpha,\beta$ are temperature scales and $\epsilon_1,\epsilon_2$ are margins.  For positive weighting, when the anchor's neighborhood is dense, the denominator of the second term gets larger and so does $W_{ii}$. As will be shown in gradient computation (Eq.~\eqref{eq:grad_hal}), a large $W_{ii}$ scales up positive sample's gradient. Analogously, for negative weighting, a dense neighborhood leads to a large $W_{it}$ and increases the gradient of that negative sample in local weighting.

\begin{table*}[!htbp]
\scriptsize
\setlength{\tabcolsep}{4pt}
\caption{Quantitative results on Flickr30k~\cite{young2014image}. ``ours'' means our own implementation. }
\label{Table:f30k}
\centering
\begin{tabular}{cllccccccccccc}
\toprule
\multirow{2}{*}{\#} &  \multirow{2}{*}{{\bf architecture}} & \multirow{2}{*}{{\bf loss}} \phantom{abc} & \multicolumn{5}{c}{\bf image$\rightarrow$text}  \phantom{abc} & \multicolumn{5}{c}{\bf text$\rightarrow$image} \\ \cmidrule(l){4-8} \cmidrule(l){9-13}
  &  & &  \textsf{R@$1$} & \textsf{R@$5$} & \textsf{R@$10$}  & \textsf{Med r}    & \textsf{Mean r} &  \textsf{R@$1$} & \textsf{R@$5$} & \textsf{R@$10$}  & \textsf{Med r} & \textsf{Mean r}  & \textsf{rsum}  \\ 
  \midrule
 
 1.1 & \multirow{3}{*}{GRU+VGG19} &  \textsc{Sum} & 30.0 & 59.6 & 67.7 & 4.0 & 34.7 & 22.8 & 49.4 & 61.4 & 6.0 & 47.5 &  291.0  \\
 1.2 & & \textsc{Max} & 30.1 & 56.3 & 67.9 & 4.0 & 30.5 & 21.3 & 47.1 & 58.7 & 6.0 & 40.2 & 281.4 \\
 1.3 & & \textsc{Hal} & \textbf{38.4} & \textbf{63.3} & \textbf{73.4}& \textbf{3.0} & \textbf{20.1} & \textbf{26.7} & \textbf{53.3} & \textbf{64.9} & \textbf{5.0}& \textbf{32.1}& \textbf{320.0} \\
 \midrule
  1.4 & \multirow{3}{*}{\shortstack[l]{ Order (VGG19, ours) \\ \cite{vendrov2015order}}} &  \textsc{Sum} & 31.4 & 58.3 & 69.4 & 4.0 & 26.9 & 24.2 & 50.9 & 62.9 & 5.0 & 34.3 & 297.1  \\
 1.5 &  & \textsc{Max} &  32.1 & 58.0 & 69.9 & 4.0 & 23.1 & 22.7 & 49.4 & 61.3 & 6.0 & 32.9& 293.4\\
 1.6 & & \textsc{Hal} & \textbf{36.4} & \textbf{62.2} & \textbf{73.0} & \textbf{3.0} & \textbf{20.4} & \textbf{26.6} & \textbf{54.4} & \textbf{65.6} & \textbf{4.0} & \textbf{31.0} & \textbf{318.3} \\
 \midrule
 1.7 &  \multirow{2}{*}{\shortstack[l]{SCAN \\ \cite{lee2018stacked}}} & \textsc{Max} & 67.9 & 89.0 & 94.4 & - & - & 43.9 & \textbf{74.2} & \textbf{82.8} & - & - & 452.2 \\
 1.8 & & \textsc{Hal} &  \textbf{68.6} & \textbf{89.9} & \textbf{94.7} & 1.0 & 3.3 & \textbf{46.0} & 74.0 & 82.3 & 2.0 & 14.3 & \textbf{455.5}\\
\bottomrule
\end{tabular}
\vspace{-0.1cm}
\end{table*}
\subsubsection{b) Local weighting through loss function.}
Here we adapt the NCA loss for classification for our context of producing a matching among two sets of points:
\begin{equation}
\begin{split}
\mathcal{L_\textsc{Hal}} =  \frac{1}{N}\sum_{i=1}^{N} \Big( \frac{1}{\gamma}\log(1+\sum_{m\not = i} e^{\gamma W_{mi}(S_{mi}-\epsilon)}) \\
+\frac{1}{\gamma}\log(1+\sum_{n\not=i} e^{\gamma W_{in}(S_{in}-\epsilon)}) \\ 
- \log (1+ W_{ii}S_{ii} )\Big), \\
\end{split}
 \label{eq:new_loss2}
\end{equation}
where $\gamma$ is a temperature scale; $\epsilon$ is a margin; $N$ is number of samples within the mini-batch. And the gradients with respect to negative and positive samples are computed as:
\begin{equation}
\begin{split}
w^+ = \bigg| \frac{\partial \mathcal{L_{\textsc{Hal}}}}{\partial {S_{ij}}^+}\bigg| = \frac{W_{ij}}{1+W_{ij}S_{ij}}\text{ if }i=j ,\\
w^- = \bigg| \frac{\partial \mathcal{L_{\textsc{Hal}}}}{\partial {S_{ij}}^-}\bigg| = \underbrace{\frac{W_{ij} e^{\gamma W_{ij}(S_{ij}-\epsilon)}}{1+\sum_{m\neq j}e^{\gamma W_{mj}(S_{mj}-\epsilon)}}}_{\text{weighted by image modality}} \\
+\underbrace{\frac{ W_{ij}e^{\gamma W_{ij}(S_{ij}-\epsilon)}}{1+\sum_{n\neq i}e^{\gamma W_{in}(S_{in}-\epsilon)}}}_{\text{weighted by text modality}}.
\end{split}
 \label{eq:grad_hal}
\end{equation}
Unlike a naive NCA aiming for classifying samples in only one direction, the first and second term of $\mathcal{L_{\textsc{Hal}}}$ punish mistakes made during searching targets among the two modalities in both directions. As shown in gradients, the sample is weighted according to its significance as a hub in both modalities.

\textbf{\textsc{Hal} vs \textsc{Max}.} As pointed out by \cite{lazaridou2015hubness}, \textsc{Max} actually implicitly mitigates the hubness problem by targeting the hardest triplet only. A hub, by definition, is a close (potentially nearest) neighbor to multiple queries and would thus be punished by \textsc{Max} for multiple times (in different batches).
\cite{lazaridou2015hubness}'s experiments also verified such theory empirically. However, it is a risky choice as the hardest sample within a mini-batch can easily be a pseudo hardest negative as analyzed above. As we would show in experiments, \textsc{Hal} prevails in a broader range of data and model configurations while \textsc{Max} only performs well on some specific circumstances where both training data and encoders are of ideal quality. Also, \textsc{Hal} is essentially leveraging more information than \textsc{Max}. In \textsc{Max}, only hub that violates margin the most gets to impose a gradient on network's parameters while \textsc{Hal} softly considers all hubs, big or small, by assigning them weights.

\section{Experiments}
\label{sec:4.4}

This section is divided into 1) Experimental Setups and 2) Main Results, where detailed configurations of experiments are introduced in 1) and comparison \& analysis of main results are in 2).

\subsection{Experimental Setups}
\label{sec:setup}

\textbf{Dataset.} We use MS-COCO \cite{lin2014microsoft} and Flickr30k \cite{young2014image} as our experimental datasets. For MS-COCO, there have been several different splitting protocols being used in the community. We use the same split as \cite{karpathy2015deep}: 113,287 images for training, 5,000 for validation and 5,000 for testing.\footnote{Note that 1 image in MS-COCO and Flickr30k has 5 captions, so 5 text-image pairs are used for every image.} During testing, scores are computed as the average of 5 folds of 1k images. As many of the previous works report test results on a 1k test set (a subset of the 5k one), we would experiment with both protocols. We refer to the 1k test set as $c1$ and the 5k test set as $c2$. Flickr30k has 30,000 images for training; 1,000 for validation; 1,000 for testing.

\textbf{Evaluation metrics.}
We use \textsf{R@$K$}s (recall at $K$), \textsf{Med r}, \textsf{Mean r} and \textsf{rsum} to evaluate the results. \textsf{R@$K$}: the ratio of ``\# of queries that the ground-truth item is ranked in top $K$'' to ``total \# of queries'' (we use $K\in\{1,5,10\}$); \textsf{Med r}: the median of the ground-truth ranking; \textsf{Mean r}: the mean of the ground-truth ranking; \textsf{rsum}: the sum of \textsf{R@$\{1,5,10\}$} for both text$\rightarrow$image and image$\rightarrow$text. \textsf{R@$K$}s and \textsf{rsum} are the higher the better while \textsf{Med r} and \textsf{Mean r} are the lower the better. We compute all metrics for both text$\rightarrow$image and image$\rightarrow$text retrieval.  During training, we follow the convention of taking the model with the maximum \textsf{rsum} on validation set as the best model for testing.

\textbf{Model and training details.}
We use $300$-$d$ word embeddings and $1024$ internal states for GRU text encoder (all randomly initialized with Xavier init.~\cite{glorot2010understanding}); all image encodings are obtained from image encoders pre-trained on ImageNet (for fair comparison, we don't finetune any image encoders); $d=1024$ for both text and image embeddings. For more details about hyperparameters and training configurations please refer to Table \ref{table:config} and code release: \url{https://github.com/hardyqr/HAL}.

\subsection{Main Results}
Here we present the major quantitative and qualitative findings with analysis regarding \textsc{Hal}'s performance, hyperparameters' choice and hubs' distributions.

\begin{table*}[h]
\scriptsize
\setlength{\tabcolsep}{4pt}
\caption{Quantitative results on MS-COCO~\cite{lin2014microsoft}. First three blocks (line 2.1-2.12) are using protocol $c2$ (5k test set); the last two blocks (line 2.13-2.24) is using $c1$ (1k test set) in convenience of comparing with results reported in previous works. \textsc{Mb} means memory bank.}
\label{Table:quant}
\centering
\begin{tabular}{lllccccccccccc}
\toprule
\multirow{2}{*}{\#} &  \multirow{2}{*}{{\bf architecture}} & \multirow{2}{*}{{\bf loss}}\phantom{abc} & \multicolumn{5}{c}{\bf image$\rightarrow$text}  \phantom{abc} & \multicolumn{5}{c}{\bf text$\rightarrow$image} & \\ \cmidrule(l){4-8} \cmidrule(l){9-13}
  & & & \textsf{R@$1$} & \textsf{R@$5$} & \textsf{R@$10$}  & \textsf{Med r} & \textsf{Mean r} & \textsf{R@$1$} & \textsf{R@$5$} & \textsf{R@$10$}  & \textsf{Med r} & \textsf{Mean r}  & \textsf{rsum}  \\ 
 \midrule 
2.1  & \multirow{4}{*}{\shortstack[l]{GRU+VGG19}}  & \textsc{Sum}  &  46.9 & 79.7 & 89.5 & 2.0 & 5.9 & 37.0 & 73.1 & 85.3 & \textbf{2.0} & 11.1 & 411.5 \\ 
2.2 & & \textsc{Max} &  51.8 & 82.1 & 90.5 & \textbf{1.0} & 5.1 & 39.0 & 73.9 & 84.7 & \textbf{2.0} & 12.0 & 421.9 \\
2.3 & & \textsc{Hal}  & 55.5 & 84.3 & 92.3 & \textbf{1.0} & 4.2 & \textbf{41.9} & 75.6 & 86.7 & \textbf{2.0} & 7.8 & 436.1  \\
2.4 & &  \textsc{Hal}+\textsc{Mb} &  \textbf{56.7} & \textbf{84.9} & \textbf{93.0} & \textbf{1.0} & \textbf{4.0} & \textbf{41.9} & \textbf{75.9} & \textbf{87.1} & \textbf{2.0} & \textbf{7.2} & \textbf{439.5} \\
\midrule
2.5  & \multirow{4}{*}{\shortstack[l]{GRU+IRv2}} & \textsc{Sum} & 50.9 & 82.7 & 92.2 & 1.4 & 4.1 & 39.5 & 75.8 & 87.2 & \textbf{2.0} & 9.4 &  428.3  \\
2.6 & & \textsc{Max} & 57.0 & 86.2 & 93.8 & \textbf{1.0} & 3.5 & 43.3 & 77.9 & 87.9 & \textbf{2.0} & 8.6 & 446.0 \\
2.7 & & \textsc{Hal} & 60.2 & 87.3 & 94.4 & \textbf{1.0} & 3.3 & 44.8 & 78.2 & 88.3 & \textbf{2.0} & 7.7 & 453.2  \\
2.8 & & \textsc{Hal}+\textsc{Mb} & \textbf{62.7} & \textbf{88.0} & \textbf{94.6} & \textbf{1.0} & \textbf{3.1} & \textbf{45.3} & \textbf{78.8} & \textbf{89.0} & \textbf{2.0} & \textbf{6.3} &  \textbf{458.5} \\
\midrule 
2.9 & \multirow{4}{*}{\shortstack[l]{GRU+ResNet152}}  &  \textsc{Sum} &   53.2 & 85.0 & 93.0 & \textbf{1.0} & 3.9 & 41.9 & 77.2 & 88.0 & \textbf{2.0} & 8.7 & 438.3 \\
2.10 &  &  \textsc{Max}  & 58.7 & 88.2 & 94.8 & \textbf{1.0} & 3.2 & 45.0 & 78.9 & 88.6 & \textbf{2.0} & 8.6 & 454.2 \\
2.11 & & \textsc{Hal} & \textbf{64.4} & 89.2 & 94.9 & \textbf{1.0} & 3.0 & 46.3 & 78.8 & 88.3 & \textbf{2.0} & 7.9 & 462.0 \\
2.12 & & \textsc{Hal}+\textsc{Mb} & 64.0 & \textbf{89.9} & \textbf{95.7} & \textbf{1.0} & \textbf{2.8} &\textbf{46.9} & \textbf{80.4} & \textbf{89.9} & \textbf{2.0} & \textbf{6.1} & \textbf{466.7}  \\
\midrule
 2.13 & \multicolumn{2}{l}{\cite{kiros2014unifying} (ours)} & 49.9 & 79.4 & 90.1 & 2.0 & 5.2 & 37.3 & 74.3 & 85.9 & \textbf{2.0} & 10.8 & 416.8 \\ 
2.14 &\multicolumn{2}{l}{\cite{vendrov2015order}} & 46.7 & - & 88.9  & 2.0 & 5.7 & 37.9 & - & 85.9 & \textbf{2.0} & 8.1 & -   \\
2.15 & \multicolumn{2}{l}{\cite{huang2017instance}}  &  53.2 & 83.1 & 91.5 & \textbf{1.0} & - & 40.7 & 75.8 & 87.4 & \textbf{2.0} & - & 431.8  \\
2.16 & \multicolumn{2}{l}{\cite{liu2017learning}} & 56.4 & 85.3  & 91.5 & - & - & 43.9 & 78.1 & 88.6 & - & - &443.8 \\
2.17 & \multicolumn{2}{l}{ \cite{you2018end} } &  56.3 & 84.4 & 92.2 & \textbf{1.0} & - & 45.7 & 81.2 & 90.6 & \textbf{2.0} & - & 450.4 \\
2.18 & \multicolumn{2}{l}{\cite{wehrmann2018bidirectional} (d=1024)}& 57.8 & 87.9 & 95.6 & \textbf{1.0} & 3.3 & 44.2 & 80.4 & 90.7 & \textbf{2.0} & \textbf{5.4} & 456.6 \\
2.19 & \multicolumn{2}{l}{\cite{faghri2018vse++}  } & 58.3 & 86.1 & 93.3 & \textbf{1.0} & - & 43.6 & 77.6 & 87.8 & \textbf{2.0} & - & 446.7 \\
2.20 & \multicolumn{2}{l}{\cite{faghri2018vse++} (ours) }& 60.5 & 89.6 & 94.9 & \textbf{1.0} & 3.1 & 46.1 & 79.5 & 88.7 & \textbf{2.0} & 8.5 & 459.3  \\
2.21 & \multicolumn{2}{l}{\cite{liu-ye-2019-strong}} & 58.3 & 89.2 & 95.4 & \textbf{1.0} & 3.1 & 45.0 & 80.4 & 89.6 & \textbf{2.0} & 7.2 & 457.9\\
2.22 &  \multicolumn{2}{l}{\cite{wu2019unified}}  & 64.3 & 89.2 & 94.8 & \textbf{1.0} & - & 48.3 & 81.7 & \textbf{91.2} & \textbf{2.0} & - & 469.5  \\
2.23 & GRU+ResNet152 + \textsc{Hal} & & 65.4 & 90.4 & 96.4 & \textbf{1.0} & 2.5 & 47.4 & 80.6 & 89.0 & \textbf{2.0} & 7.3 & 469.2 \\
2.24 & GRU+ResNet152 + \textsc{Hal} + \textsc{Mb} & & \textbf{66.3} & \textbf{91.7} & \textbf{97.0} & \textbf{1.0} & \textbf{2.4} & \textbf{48.7} & \textbf{82.1} & 90.8 & \textbf{2.0}& 5.6 & \textbf{476.6} \\
\midrule
2.25 & \multicolumn{2}{l}{\cite{lee2018stacked} (t-i AVG)} & 70.9 & 94.5 & 97.8 & - & - & 56.4 & \textbf{87.0} & \textbf{93.9} & - & - &  500.5 \\
2.26 & \multicolumn{2}{l}{\cite{lee2018stacked} (t-i AVG) + \textsc{Hal}}& \textbf{78.3} & \textbf{96.3} & \textbf{98.5} & 1.0 & 2.6 & \textbf{60.1} & 86.7 & 92.8 & 1.0 & 5.8 & \textbf{512.7}\\
\bottomrule
\end{tabular}
\vspace{-0.1cm}
\end{table*}

\textbf{Comparing \textsc{Hal}, \textsc{Sum} and \textsc{Max}.}
Table \ref{Table:f30k} and \ref{Table:quant} present our quantitative results on Flickr30k and MS-COCO respectively. 
On Flickr30k, we experiment three models and \textsc{Hal} achieves significantly better performance than \textsc{Max} and \textsc{Sum} on the first two configurations.\footnote{We do not include \textsc{Hal+Mb} for \cite{vendrov2015order} as it demands GPU memory exceeding 11GB, which is the limit of our used GTX 2080Ti. Same reason applies to SCAN+\textsc{Hal}+\textsc{Mb}.} On MS-COCO, \textsc{Hal} also beats both triplet loss functions. Interestingly, while \textsc{Max} fails badly on Flickr30k, it becomes very competitive on MS-COCO. This serves as an evidence of \textsc{Max} easily overfitting to small datasets.\footnote{\cite{faghri2018vse++} showed that data augmentation techniques like random crop applied on input images can improve \textsc{Max}'s performance over small datasets.} In conclusion, \textsc{Hal} maintains its edge over \textsc{Max} and \textsc{Sum} across regardless of data and architecture configurations. Even without global weighting (memory bank), \textsc{Hal} still beats the two triplet losses by a large margin. The equipment of memory bank can usually further boosts \textsf{rsum} by another $3-5$. Also, it is worth noticing that \textsc{Hal} converges significantly faster than \textsc{Max} and \textsc{Sum}. \textsc{Hal} stabilizes after approximately $5$ epochs while \textsc{Max} and \textsc{Sum} take roughly $10$ epochs.

\begin{figure}[h]
    \centering
    \includegraphics[width=7cm]{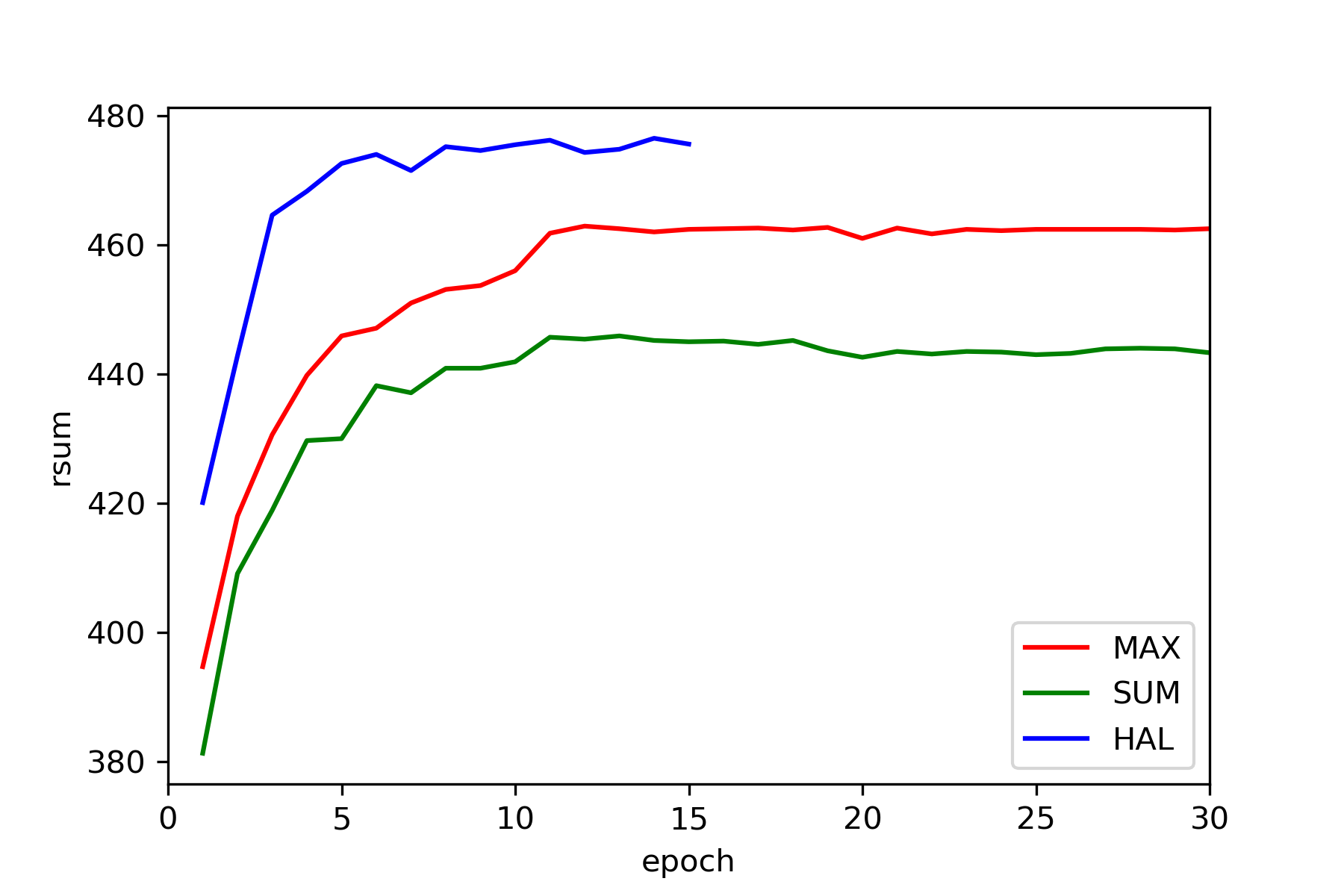}
    \caption{\small{Plotting epoch against \textsf{rsum} on validation set for comparing convergence time. All models are using GRU+ResNet152, trained \& validated on MS-COCO.}}
    \label{fig:hal_bs}
\end{figure}

\textbf{\textsc{Hal} vs. State-of-the-art.} Table~\ref{Table:quant} line 2.13-2.24 list quantitative results of both our proposed method (2.23, 2.24) and numbers reported in previous works (2.13-2.22). For fair comparison with \cite{faghri2018vse++}, we only use routine encoder architectures (GRU+ResNet152). Unlike \cite{shi2018learning,wu2019unified}, we also do not bring in any extra information to help training. With a trivial configuration of model \& data, our method is still ahead of the state-of-the-art on MS-COCO \cite{wu2019unified} by a decent margin for most metrics. Notice that we are comparing against works that use frozen image encoder (as we do). For the ones that finetuning image features, better performance is achievable \cite{song2019polysemous,shuster2019engaging}. In Table \ref{Table:quant} line 2.25, 2.26, we list SCAN \cite{lee2018stacked} alone as it incorporates additional knowledge, i.e. bottom-up attention information, from a Faster R-CNN \cite{ren2015faster} to refine the visual-semantic alignment. With such prior, it is a well-established state-of-the-art on the Text-Image matching task, having much higher \textsf{rsum}s than previous works. For SCAN, we pick configurations with the best \textsf{rsum}s on both MS-COCO and Flikr30k, switching its learning objective from \textsc{Max} to \textsc{Hal}.\footnote{An ensemble model is able to achieve even higher \textsf{rsum} but for clear comparison we do not discuss the ensemble case.} On Flikr30k, \textsc{Max} and \textsc{Hal} deliver comparable results. On MS-COCO, \textsc{Hal} is significantly stronger - \textsf{rsum} is further improved to \textbf{512.7} with \textsf{R@}$1$ improved by 7.4 and 3.7 for image$\rightarrow$text and text$\rightarrow$image respectively. We did not experiment with \textsc{Mb} due to GPU memory limits.

\textbf{The impact of batch size.} In contrast to loss functions that treat each sample equivalently, batch size does matter to \textsc{Hal} as it defines the neighborhood size where relative similarity is considered during local weighting. And \textsc{Hal} does benefit from a larger batch size as it means an expanded neighborhood. As suggested in Figure~\ref{fig:hal_bs}, on MS-COCO, \textsc{Hal} reaches a maximum \textsf{rsum} with a batch size of $512$. Note that in the NCA context, batch size is a relative concept. For Flickr30k, which is only of roughly $\frac{1}{4}$ the size of MS-COCO, we maintain the original batch size of $128$ to cover roughly the same range of neighborhood.

\begin{figure}[h]
    \centering
    \includegraphics[width=7cm]{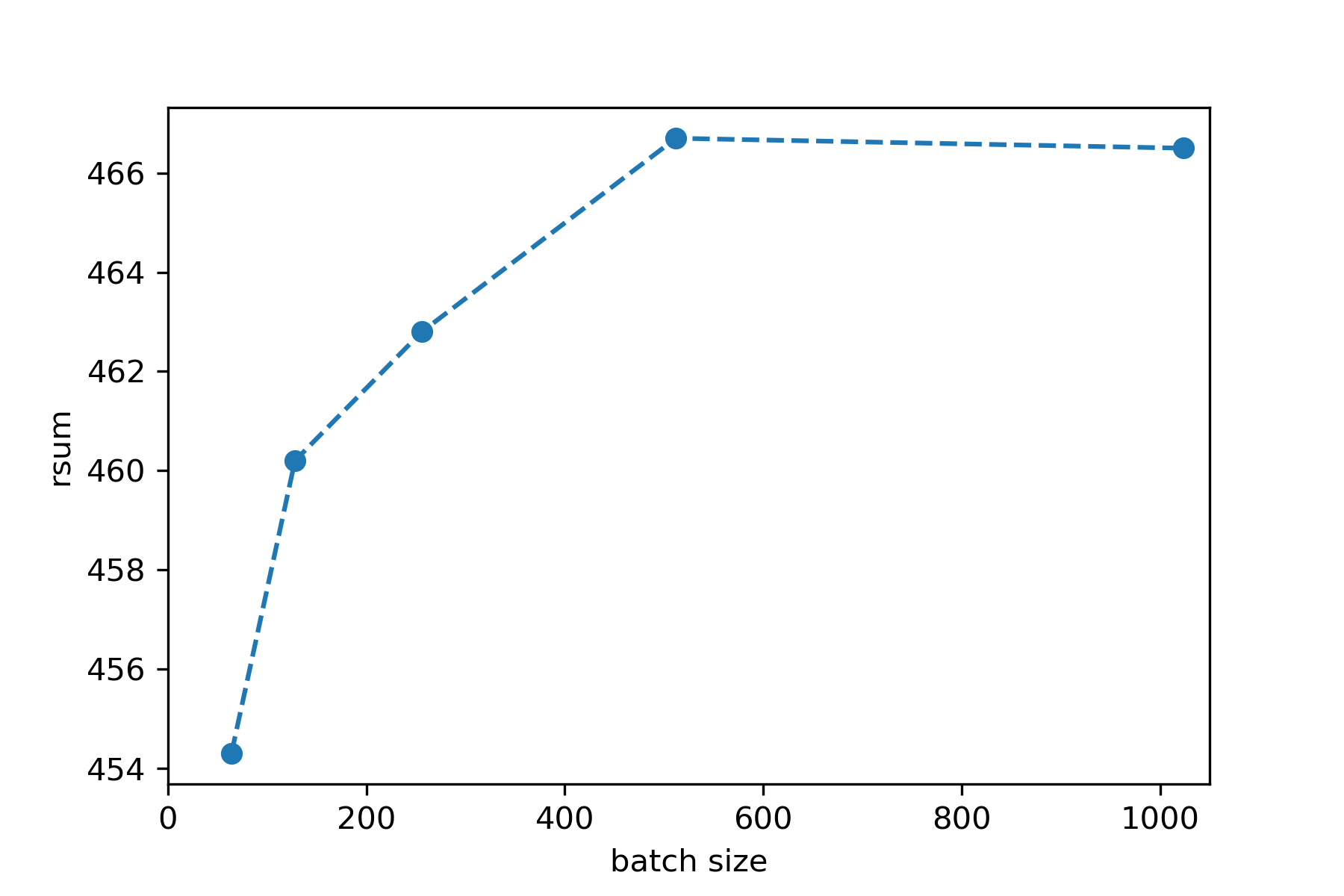}
    \caption{\small{Plotting batch size used by \textsc{Hal} against \textsf{rsum}. All models are using GRU+ResNet152, trained \& tested on MS-COCO $c2$.}}
    \label{fig:hal_bs}
\end{figure}

\begin{figure}[h]
    \centering
    \includegraphics[width=7cm]{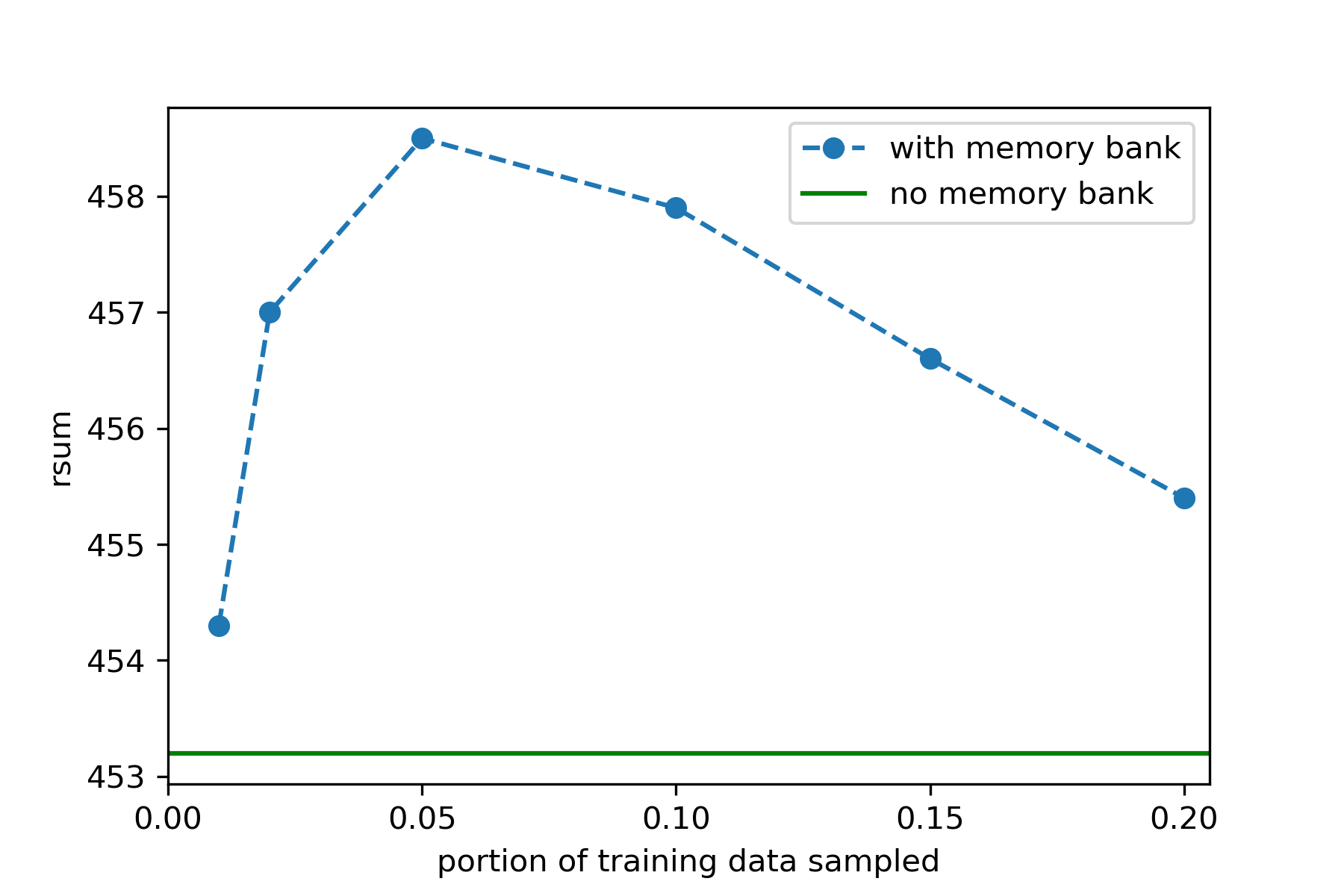}
    \caption{\small{Plotting \textsf{rsum} against \textsc{Hal}'s memory bank size. \textsc{Hal} without memory bank is also provided as a baseline. All data points are produced with GRU+IRv2 as the base model and are trained \& tested on MS-COCO $c2$.}}
    \label{fig:hal_mb_size}
\end{figure}

\textbf{The impact of size of memory bank.} The \textsc{Mb} in \textsc{Hal} has two hyperparameters: 1) $k$, which characterizes the scope of neighborhood being considered for global statistics, and 2) memory bank's size. Their relative scales matter for mining informative samples in the top-k neighborhood. When $k$ is fixed, we search the most appropriate memory bank size and find that $5\%$ of training data is ideal as suggested in Figure \ref{fig:hal_mb_size}. The top-k neighborhood of a too large memory bank might be filled with noisy samples (potentially being incorrectly labeled).

\begin{table*}[!htbp]
\small
\setlength{\tabcolsep}{4pt}
\caption{Experiment configurations. }
\label{table:config}
\centering
\begin{tabular}{cllccccccccccc}
\toprule
\textbf{\#} & \textbf{Datasets} & \textbf{models} & \textbf{hyperparameters} \\ 
  \midrule
 3.1 & \multirow{6}{*}{MS-COCO} &   2.1, 2.5, 2.9, 2.13  &  margin=0.2, lr=0.001, lr\_update=10, bs=128, epoch=30 \\
 3.2 & & 2.2, 2.6, 2.10, 2.20&  margin=0.2, lr=0.0002, lr\_update=10, bs=128, epoch=30  \\
 3.3 & & 2.11, 2.23 & $\gamma$=30, $\epsilon$=0.3, lr=0.001, lr\_update=10, bs=512, epoch=15\\
 
 \multirow{2}{*}{3.4} & & \multirow{2}{*}{2.12, 2.24} & $\gamma$=30, $\epsilon$=0.3, $\alpha$=40, $\beta$=40, $\epsilon_{1}$=0.2, $\epsilon_{2}$=0.1\\
 & & & lr=0.001, lr\_update=10, bs=512, epoch=15\\

 3.5 & & 2.26 & $\gamma$=100,\ $\epsilon$=1.0, lr=0.0005, lr\_update=10, bs=256, epoch=20\\
 \midrule
  3.6 & \multirow{4}{*}{\shortstack[l]{Flickr30k}} & 1.1, 1.4 & margin=0.05, lr=0.001, lr\_update=10, bs=128, epoch=30\\
  3.7 & & 1.2, 1.5 & margin=0.05, lr=0.0002, lr\_update=15, bs=128, epoch=30\\
  3.8 & & 1.3 & $\gamma$=60, $\epsilon$=0.7, lr=0.001, lr\_update=10, bs=128, epoch=15\\
  3.9 & & 1.8 & $\gamma$=70,\ $\epsilon$=0.6, lr=0.0005, lr\_update=10, bs=128, epoch=30\\

\bottomrule
\end{tabular}
\vspace{-0.1cm}
\end{table*}

\section{Related Work}
In this section, we introduce works from three fields that are highly-related to our work: 1) text-image matching and VSE; 2) deep metric learning; 3) tackling the hubness problem in various contexts.

\subsubsection{Text-image Matching and VSE.}
 Since the dawn of deep learning, works have emerged using a two-branch architecture to connect language and vision. \citet{weston2010large} trained a \emph{shallow} neural network to map word-image pairs into a joint space for image annotation. In 2013, \citet{frome2013devise} brought up the term VSE and trained joint embeddings for sentence-image pairs. Later works extended VSE for the task of text-image matching \cite{hodosh2013framing,kiros2014unifying,gong2014improving,vendrov2015order,tsai2017learning,faghri2018vse++,wang2019learning}, which is also our task of interest. Notice that text-image matching is different from generating novel captions for images \cite{lebret2015phrase,karpathy2015deep} but is to retrieve existing descriptive texts or images in a database.
 
 While many of these works improve model architectures for training VSE, few have tackled the shortcomings in learning objectives. \citet{faghri2018vse++} made the latest attempt to reform the long being used \textsc{Sum} loss. Their proposed \textsc{Max} loss is indeed a much stronger baseline than \textsc{Sum} in most data and model configurations. But it fails significantly when the dataset is small or noise is contained. \citet{liu-ye-2019-strong} eased such deficiency by relaxing \textsc{Max} into a top-K triplet loss. \citet{shekhar2017foil,shi2018learning} raised similar concerns. They mainly focused on creating better training data while we target the training objective itself.
 
 
\subsubsection{Deep Metric Learning.}
Text-image matching is an open-set task where matching results are identified based on similarity of pairs, instead of assigning probabilities to specific labels in a closed set. Such property coincides with the idea of metric learning, which utilizes relative similarities among pairs to cluster samples of same class in embedding space. Entering the deep learning age, deep neural net based metric learning is widely applied in various tasks including image retrieval \cite{oh2016deep,wang2019multi}, face recognition \cite{schroff2015facenet}, person re-identification \cite{binomiance}, etc.. We use kindred philosophy in our context of matching two sets of data points. Works on deep metric learning that inspired our model are discussed here.

Neighborhood Component Analysis (NCA) \cite{goldberger2005neighbourhood} introduced the foundational philosophy for metric learning where a stochastic variant of K-Nearest-Neighbor score is directly maximized. \cite{binomiance,oh2016deep,sohn2016improved,wang2019multi} further developed the idea, leveraging the gradient of NCA-based loss to discriminatively learn from samples of different importance. \cite{wu2018improving} proposed a method that computes only part of NCA-based loss's gradient, so that NCA on a large scale is computationally feasible.

\subsubsection{Tackling the Hubness Problem.}
\label{sec:hubness}
We have stated what the hubness problem is in the introduction. Now we introduce several efforts tackling the hubness problem in various contexts. \cite{zhang2017learning} pointed out the wide existence of hubs in text-image embeddings but did not address them. Though not receiving enough attention in VSE literature, hubness problem has recently been extensively explored in Bilingual Lexicon Induction (BLI). BLI is the task of inducing word translations from monolingual corpora in two languages~\cite{irvine2017comprehensive}. In terms of finding correspondence between two sets of vectors, it is analogous to our task of interest. \cite{smith2017offline,lample2018word} proposed to first conduct a direct Procrustes Analysis and then use criteria that heavily punish hubs during inference to reduce the hubness problem. While it is indeed efficient in finding a better matching, the actual quality of embedding is not improved. \citet{joulin2018loss} integrated the inference criterion \textsc{Csls} from \cite{lample2018word} into a least-square loss and trained a transformation matrix end-to-end to mitigate hubness problem. Though this work has a similar philosophy to ours, it is specifically designed for BLI and only trains one linear layer over two sets of word vectors. When \textsc{Csls} is appended to a triplet loss, it is merely a resampling of hard samples, making it non-special in terms of both form and intuition.

\section{Conclusion}
We introduce a novel loss \textsc{Hal} for mitigating visual semantic hubs during training text-image matching models. The self-adaptive loss \textsc{Hal} leverages the inherit nature of Neighborhood Component Analysis (NCA) to identify information of hubs, from both a global and local perspective, giving considerations to robustness and hard sample mining at the same time. Our method beats two prevalent triplet-based objectives across different datasets and model architectures by large margins. Though our methods have only experimented on the task of text-image matching, there remains to be other cross-modal mapping tasks requiring obtaining a matching, e.g. content-based image retrieval, document retrieval, document semantic relevance, Bilingual Lexicon Induction, etc.. \textsc{Hal} can presumably be used in such settings as well.

\section{Acknowledgments}
We thank anonymous reviewers for their careful feedbacks, based on which we were able to enhance the work. We thank our family members for unconditionally supporting our independent research.  The author Fangyu Liu gives special thanks to 1) Prof. Lili Mou, who voluntarily spent time reading and discussing the rough ideas with him at the very beginning; 2) his aunt Qiu Wang who supplied him with GPU machines; 3) his labmates Yi Zhu and Qianchu Liu from Language Technology Lab for proofreading the camera-ready version.

\bibliography{ref}
\bibliographystyle{aaai}

\end{document}